\documentclass[manuscript,screen]{acmart}
\AtBeginDocument{%
  }

\usepackage[acronym]{glossaries}
\usepackage{bm}
\usepackage{amsmath,amsfonts}
\usepackage{dsfont}
\usepackage{siunitx}

\usepackage{xr}
\makeglossaries
\newacronym{av}{AV}{Autonomous vehicle}
\newacronym{ade}{ADE}{Average Displacement Error}
\newacronym{fde}{FDE}{Final Displacement Error}
\newacronym{pgd}{PGD}{Projected Gradient Descent}
\begin{document}

\title{Realistic Adversarial Attacks for Robustness Evaluation of Trajectory Prediction Models via Future State Perturbation}

\author{Julian F. Schumann}
\authornote{Both authors contributed equally to this research.}
\email{J.F.Schumann@tudelft.nl}
\orcid{0000-0002-4482-5122}
\author{Jeroen Hagenus}
\authornotemark[1]
\email{J.Hagenus@student.tudelft.nl}
\affiliation{%
  \institution{Department of Cognitive Robotics, TU Delft}
  \city{Delft}
  \country{Netherlands}
}

\author{Frederik Baymler Mathiesen}
\email{F.B.Mathiesen@tudelft.nl}
\orcid{0000-0002-2243-0445}
\affiliation{%
  \institution{Delft Center for Systems and Control, TU Delft}
  \city{Delft}
  \country{Netherlands}
}

\author{Arkady Zgonnikov}
\email{A.Zgonnikov@tudelft.nl}
\orcid{0000-0002-6593-6948}
\affiliation{%
  \institution{Department of Cognitive Robotics, TU Delft}
  \city{Delft}
  \country{Netherlands}
}

\renewcommand{\shortauthors}{Schumann et al.}

\begin{abstract}
    Trajectory prediction is a key element of autonomous vehicle systems, enabling them to anticipate and react to the movements of other road users. Evaluating the robustness of prediction models against adversarial attacks is essential to ensure their reliability in real-world traffic. However, current approaches tend to focus on perturbing the past positions of surrounding agents, which can generate unrealistic scenarios and overlook critical vulnerabilities. This limitation may result in overly optimistic assessments of model performance in real-world conditions.  
    In this work, we demonstrate that perturbing not just past but also future states of adversarial agents can uncover previously undetected weaknesses and thereby provide a more rigorous evaluation of model robustness. Our novel approach incorporates dynamic constraints and preserves tactical behaviors, enabling more effective and realistic adversarial attacks. We introduce new performance measures to assess the realism and impact of these adversarial trajectories. Testing our method on a state-of-the-art prediction model revealed significant increases in prediction errors and collision rates under adversarial conditions. Qualitative analysis further showed that our attacks can expose critical weaknesses, such as the inability of the model to detect potential collisions in what appear to be safe predictions. These results underscore the need for more comprehensive adversarial testing to better evaluate and improve the reliability of trajectory prediction models for autonomous vehicles. 
\end{abstract}




\maketitle

\section{Introduction}\label{ch:introduction}

\begin{figure*}
    \centering
    \includegraphics[width = \textwidth]{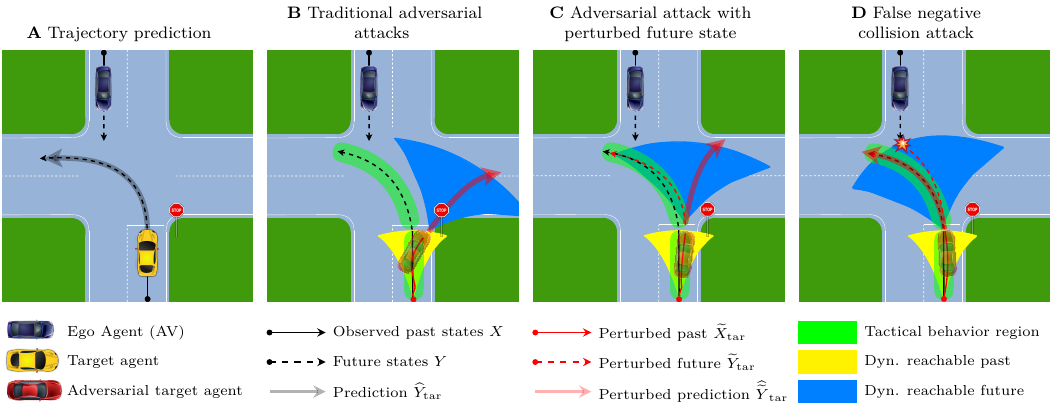}
    \caption{Scenarios of adversarial attacks by the target agent on trajectory prediction model of the ego agent (AV). \textbf{A)} In the baseline, non-adversarial scenario, based on the observed past trajectory $X = \{X_{\text{ego}}, X_{\text{tar}}\}$ of the target agent, the ego agent makes a prediction $\widehat{Y}$ that aims to closely capture the (unknown) future states $Y$. \textbf{B)} Adversarial (target) agent aims to attack the trajectory prediction model of the AV with the goal for that model to produce an erroneous prediction of the target agent. Existing approaches in the literature typically generate adversarial attacks by perturbing past states of the target vehicle $\widetilde{X}$ in such a way that the corresponding prediction $\widehat{\widetilde{Y}}$ deviates as much as possible from the ground-truth future states $Y$. However, existing methods do not guarantee that these future states $Y$ actually lie in the reachable set of the target agent after the perturbation, which can result in an overall unsuccessful attack where $\widehat{\widetilde{Y}}$ is actually consistent with the actual future of the adversarial agent. \textbf{C)} In our proposed approach, we generate (and constrain) perturbations not only for the past states $\widetilde{X}$ but also future states $\widetilde{Y}$ of the target agent. This ensures that the observed future $Y$ stays dynamically feasible, which ensures that prediction errors resulting from the perturbation are meaningful. \textbf{D)} Our approach allows us to generate a novel type of adversarial attacks: \textit{false negative collision attack}, where the adversarial agent takes a future trajectory that collides with the AV but deceives the AV into predicting a non-colliding trajectory.  In B) - D), the yellow area represent the dynamically reachable set starting from the first observation, while the blue region represents the same starting from the prediction point. The green region represents a simple distance constraint often employed in adversarial attacks in the literature.}
    \label{Fig:Introduction_image}
\end{figure*}

Trajectory prediction -- the task of forecasting the future positions of traffic participants over time -- is a critical component of autonomous vehicle (\acrshort{av}) systems, allowing them to adapt to the actions of nearby road users (Figure \ref{Fig:Introduction_image}A). As \acrshortpl{av} become more prevalent, ensuring the robustness and reliability of the prediction models is essential for public safety \cite{hagenus2024survey,llorca2024testing}. The use of adversarial trajectories -- specifically generated to deceive the model prediction as much as possible -- has emerged as a method for evaluating robustness of trajectory prediction models, helping to uncover vulnerabilities that could lead to dangerous situations on the road, particularly in the presence of adversarial agents \cite{cao_advdo_2022, saadatnejad_are_2022, cao_robust_2023, zhang2022adversarial, jiao_semi-supervised_2023, tan_targeted_2022}. The ability of a model to withstand such attacks is known as adversarial robustness \cite{tocchetti2022ai}.

Recent advances in adversarial attack generation have shown promise in exposing system vulnerabilities \cite{zhang2022adversarial, cao_advdo_2022, yin2024saattack, tan_targeted_2022, fan2024adversarial}. Existing attack generation methods have traditionally focused on perturbing the observed position of the target vehicle. These methods can create scenarios that challenge trajectory prediction models, leading to prediction errors or simulated collisions (see Section~\ref{sec:related_work}). State-of-the-art approaches demonstrate two key capabilities. First, they generate trajectories that are dynamically feasible, ensuring that attacks can actually be reproduced in a real-world setting \cite{zhang2022adversarial}. Second, the generated trajectories adhere to traffic norms, ensuring that resulting attacks cannot be trivially detected \cite{cao_advdo_2022,zhang2022adversarial}, and ensuring that evaluations of prediction models on the perturbed data are meaningful.
However, a key limitation of existing methods is their exclusive focus on perturbing \textit{observed} (past) states of the target vehicle, disregarding the effect of perturbations on the vehicle's future states. This means that existing attack methods can result in adversarial trajectories that restrict the agent to a different \textit{tactical behavior} (maneuver) than the original trajectory, potentially negating the real-life effect of the attack.

As an example, consider an intersection scenario where an adversarial agent intends to turn left, but aims to deceive the ego vehicle (AV across the intersection) into believing that it would \textit{not} turn left (Figure~\ref{Fig:Introduction_image}B). If the future states of the adversarial agent are not taken into account when generating the attack (as is the case for existing methods), the perturbation of the past trajectory might put the agent in the state from which the left turn is kinematically infeasible. As a result, the adversarial prediction would be consistent with the actual behavior of the agent, meaning that the attack would not actually deceive the ego agent (Figure~\ref{Fig:Introduction_image}B). 

In summary, disregarding future states when generating trajectory perturbations limits the ability of existing attack methods to represent adversarial behaviors that \acrshort{av}s could encounter in the real world. This in turn limits the effectiveness of adversarial testing of trajectory prediction models. The challenge thus lies in generating adversarial trajectories that not only deceive  trajectory prediction models, but also remain consistent with the target agent's original tactical behavior throughout the entire trajectory, i.e., for both past and future states.

In this paper, we present a novel method for generating adversarial attacks that accounts for both observed and future states. Our approach improves on existing techniques by introducing constraints on future states, ensuring that the entire adversarial trajectory maintains realistic driving behavior (Figure \ref{Fig:Introduction_image}C). This allows us to create more subtle and potentially dangerous attack scenarios, such as slight modifications to the path of a vehicle that lead to unexpected collisions while appearing benign to the prediction model (Figure \ref{Fig:Introduction_image}D). By considering future states during attack generation, we ensure that adversarial trajectories remain plausible and reflective of real-world driving scenarios, leading to a more comprehensive and realistic evaluation of the robustness of trajectory prediction models. The contributions of this paper can be summarized as follows:

\begin{itemize}
    \item We introduce a new method to generate dynamically feasible adversarial attacks that are consistent with the observed tactical behavior, allowing for a more meaningful evaluation of the robustness of a prediction model. Building on the approach by Cao et al.~\cite{cao_advdo_2022} to perturb not the trajectory in itself, but rather extracted control inputs of a dynamical model, we extend this approach to cases where only position data is available, while considering the agents ability to reverse. Importantly, we increase the meaningfulness of such attacks by constraining the planned perturbed trajectory as well. 
    \item We propose a new type of adversarial attack, the (\textit{false negative collision attack}). This goes beyond that standard approach of maximizing the prediction error, and instead aims to cause a collision by fooling the ego agent into discarding such a possibility until it is too late.
    \item Novel measures to assess the effectiveness and realism of adversarial attacks on trajectory prediction models. The measures are based on the distance between ground truth and perturbed past trajectory as well as the amplitude of the control actions required to execute the perturbed trajectory. 
\end{itemize}

\section{Related work}
\label{sec:related_work}
\subsection{Adversarial attack generation methods}
Existing methods of adversarial attack generation for trajectory prediction models almost exclusively generate adversarial scenarios by perturbing real traffic scenarios \cite{cao_advdo_2022, yin2024saattack, tan_targeted_2022, zhang2022adversarial, wang_advsim_2021} (see however~\cite{rempe2022generating} for an approach using generative models to sample adversarial scenarios or~\cite{pourkeshavarz2024adversarial} for backdoor attacks on a models training data). Perturbation-based attack generation methods aim to find the ``worst'' possible perturbation for a given scenario and trajectory prediction model. They can be divided into black-box attacks, which only use the output of the model, and white-box attacks, which use gradient information to find perturbations.

Black-box attacks like particle swarm optimization~\cite{zhang2022adversarial}, do not require model gradient information but can be slow and unreliable~\cite{cao_advdo_2022}. In contrast, white-box attacks, using some form of gradient descent \cite{cao_advdo_2022, zhang2022adversarial, yin2024saattack, tan_targeted_2022}, are generally effective but require the adversarial agent to have access to the model gradient.
In cases where access to gradients is restricted, for example due to intellectual property concerns, this challenge can be addressed by using surrogate models and generating transferable attacks \cite{qin2023training}. While our work only focuses on utilizing a white-box approach, all those methods could be used to pursue varying attack objectives. 

\subsection{Adversarial attack objectives}
Creating adversarial attacks involves specifying the attack objective for robustness evaluation, with different lines of attack being proposed in the literature. 

\begin{itemize}
    \item \textbf{Average or final displacement error (\acrshort{ade}/\acrshort{fde}) attack}:
    Maximize either the average or final distance between the observed ground truth and the prediction made by the behavior model based on the perturbed past~\cite{cao_advdo_2022, jiao_semi-supervised_2023, zhang2022adversarial, duan_causal_2023,zhang2022adversarial,saadatnejad_are_2022}. 
    
    \item \textbf{Lateral/longitudinal attack}. 
    Maximize the displacement projected either along the trajectory (longitudinal) or perpendicular to it (lateral), to falsely suggest behaviors such as large accelerations or lane changes~\cite{jiao_semi-supervised_2023, duan_causal_2023, zhang2022adversarial, dong2025safe}.
    
    \item \textbf{Targeted attack}. Minimize the distance between the prediction and a user-defined target trajectory~\cite{tan_targeted_2022}.
    While this method allows for manipulation of the output of the model, the range of possible outcomes is often more constrained compared to other attack methods.
    
    \item \textbf{Collision attack}. Minimize the distance between the prediction and the ground truth trajectory of the ego vehicle, to deceive the model into predicting a collision~\cite{wang_advsim_2021, saadatnejad_are_2022, dong2025safe}. 
\end{itemize} 

\subsection{Constraints used during adversarial attacks}
As simple attack objectives like ADE/FDE could be maximized by using very large (and therefore unrealistic and easily detectable) perturbations, previous work generally adopted the approach of limiting the perturbation magnitude to maintain the dynamic feasibility of the attacks. Two approaches to constraining perturbation magnitude have been used in the literature:
\begin{itemize}
    \item \textbf{Constraining vehicle positions}: Zhang et al.~\cite{zhang2022adversarial} use a method referred to as a ''Search attack'' to constrain trajectories by rescaling an initial perturbation so that velocity, accelerations and jerk lie within fixed bounds based upon the spread of those properties in the original dataset. Yin et al.~\cite{yin2024saattack} enhanced this approach by substituting the data statistics method with a continuous curvature model called an ''SA-attack'' using a pure pursuit method (a path-tracking algorithm that generates a trajectory by continuously pursuing a point on the path at a fixed look-ahead distance from the current position \cite{coulter1992implementation}). Namely, a dynamic model is fit between initial perturbations to generate trajectories that ensure dynamic feasibility at specific speeds. As an alternative approach, Cao et al.~\cite{cao_advdo_2022} did not use hard constraints, but instead added a penalty to the attack objective for overly large perturbations.
    
    \item \textbf{Constraining vehicle control actions}: Cao et al.~\cite{cao_advdo_2022} proposed generating adversarial trajectories by perturbing control actions (with hard constraints) derived from the agent's spatial position, heading angle, and velocity using a kinematic bicycle model. This ensures both that the trajectories adhere to the physical constraints associated with a specific vehicle and that the perturbed control actions remain consistent with those of the unperturbed trajectory. Meanwhile, Dong et al.~\cite{dong2025safe} added a regularization term to the attack objective to minimize curvature and its derivative in the perturbed trajectories.
\end{itemize}

Fusing these two approaches, Fan et al.~\cite{fan2024adversarial} used the discriminator of a generative adversarial network (GAN) to keep the perturbed trajectories similar to the unperturbed one, after which optimized the control inputs of bicycle model to create a kinematically feasible trajectory optimized to match the generated one as closely as possible.

Our proposed method could generate attacks by perturbing both spatial positions \cite{zhang2022adversarial} or control actions \cite{cao_advdo_2022}, however, we will focus on the more novel, latter approach in the following sections. 

\section{Background}
\subsection{Trajectory prediction}
The goal of trajectory prediction is to forecast the future movements of an traffic agent $i$ based on their past observed trajectory $X_i = \left\{\bm{x}_{i}(t) \mid t \in \{-H+1, \ldots, 0\} \right\}$, that contains $H$ positions with a time step size $\Delta t$. Similarly, we can also collect the future trajectory of the agent $Y_i = \left\{\bm{y}_{i}(t) \mid t \in \{1, \ldots, T\} \right\}$ with the prediction horizon $T$. In one scene, $X = \{X_i , \ldots, X_N\}$ is the combined history for all $N$ agents in the current scene (similarly, $Y$ is the combined future).

A probabilistic model $\mathds{P}_{\bm{\theta}}$ with trainable parameters $\bm{\theta}$ then generates $K$ equally likely trajectories $\widehat{Y}^k \sim \mathds{P}_{\bm{\theta}}(\cdot|X)$ with $\widehat{Y}=\{\widehat{Y}^k \mid k \in \{1, \ldots, K\} \}$ from the same initial state $X$. Generally, such models are trained to minimize the distance of those trajectories to the ground truth observation $Y$, or maximize the likelihood $\mathds{P}_{\bm{\theta}}(Y|X)$ of $Y$.

While there are models that actually predict the trajectories of all agents simultaneously~\cite{yuan_agentformer_2021,girgis2021latent,rowe2023fjmp,aydemir2023adapt}, most models predict agents independently, that is, they have the form $\mathds{P}\left(Y_i\mid X_i, X \setminus X_i \right)$.

\subsection{Adversarial attacks} 
In the case of an adversarial attack, an adversarial agent, also referred to as a target agent with trajectory $X_{\text{tar}}$, attempts to deceive the prediction model employed by the autonomous vehicle (ego agent $X_{\text{ego}}$) into making incorrect predictions (Figure \ref{Fig:Introduction_image}B). 

As discussed in Section~\ref{sec:related_work}, most existing attack approaches generate adversarial trajectories by modifying an existing non-adversarial trajectory. Specifically, the observed (past) states of the target agent are modified using perturbations $\delta_{X}$, resulting in the perturbed past trajectory $\widetilde{X}_{\text{tar}} = X_{\text{tar}} + \delta_{X}$. The attacked model then makes the (generally probabilistic) predictions $\widehat{\widetilde{Y}}_{\text{tar}} \sim \mathds{P}_\theta(\cdot|\widetilde{X}_{\text{tar}}, X_{\text{ego}})$, which are then used to evaluate the robustness of the models.

\subsection{Projected Gradient Descent} \label{ch:PGD_problem_definition}
To generate adversarial attacks using a white-box attack strategy, we use Projected Gradient Descent (\acrshort{pgd}). \acrshort{pgd} is an extension of traditional gradient descent methods for constrained optimization problems, iteratively refining the perturbations using alternating gradient and projection steps \cite{schmidt2020projected}. Given an initial perturbation $\delta^0 = \bm{0}$, we perform two steps in each iteration $m \in \{1, 2, ..., M_{max}\}$. First, a gradient step based on the loss function $\mathcal{L}$ utilizes a step size $\alpha$ to update the perturbation $\delta^{m-1}$, producing an intermediate perturbation $\widehat{\delta}^{m-1}$:
\begin{equation}
    \widehat{\delta}^{m-1} = \; \delta^{m-1} - \alpha^{m-1} \nabla_{\delta} \mathcal{L}\label{eq:PGD_1}
\end{equation}
Second, a projection step finds an updated perturbation $\delta^{m}$ by projecting $\widehat{\delta}^{m-1}$ onto the feasible set $C$ (Equation \ref{eq:PGD_2}).
\begin{equation} 
    \delta^{m} = \; \underset{v \in C}{\mathrm{argmin}} \left\| v - \widehat{\delta}^{m-1} \right\| \label{eq:PGD_2}
\end{equation}

\subsection{Adversarial attack objectives} \label{ch:attack_functions_PGD}
In this paper, we assume that the loss function $\mathcal{L}$ of the attacked model can be a sum of multiple objective or constraint functions $l$. The following attack objectives $l$ are common in the literature:

\begin{itemize}
     \item \textbf{ADE/FDE attack} tries to maximize the average or final distance between ground truth $Y$ and adversarial predictions $\widehat{\widetilde{Y}}$~\cite{cao_advdo_2022, jiao_semi-supervised_2023, zhang2022adversarial, duan_causal_2023}.
     \begin{equation}\label{eq:displacement_simple}
        \begin{aligned}
            l_{\text{ADE}}({Y}_{\text{tar}},\widehat{\widetilde{Y}}_{\text{tar}}) &= -\frac{1}{K T} \sum_{k=1}^{K} \sum_{t=1}^{T}  \left\Vert \widehat{\widetilde{\bm{y}}}_{\text{tar}}^k(t) - \bm{y}_{\text{tar}}(t) \right\Vert \\
            l_{\text{FDE}}({Y}_{\text{tar}},\widehat{\widetilde{Y}}_{\text{tar}}) &= -\frac{1}{K}  \sum_{k=1}^{K} \left\Vert \widehat{\widetilde{\bm{y}}}_{\text{tar}}^k(T) - \bm{y}_{\text{tar}}(T) \right\Vert
        \end{aligned}
    \end{equation}

    \item \textbf{Collision attack} aims to mislead the ego agent into predicting a collision (i.e., a false positive collision prediction) by minimizing the distance between the predicted positions of the target agent $\widehat{\widetilde{Y}}_{\text{tar}}$ and the ground truth future positions of the ego agent $Y_{\text{ego}}$~\cite{wang_advsim_2021, saadatnejad_are_2022}:
    \begin{equation}\label{eq:collision_simple}
        l_{\text{Col}}(Y_{\text{ego}},\widehat{\widetilde{Y}}_{\text{tar}}) = \frac{1}{K}  \sum_{k=1}^{K} \min_{t \in \{1, \ldots, T\}} \left\Vert \widehat{\widetilde{\bm{y}}}_{\text{tar}}^k(t) - \bm{y}_{\text{ego}}(t) \right\Vert
    \end{equation}
\end{itemize}

\section{Method} \label{ch:proposed_strategies}
\subsection{Perturbation strategy}
To improve on the realism of the traditional approach of perturbing the past states of the target vehicle $X_{\text{tar}}$, in our method we also perturb the future states $Y_{\text{tar}}$ with perturbation $\delta_{Y}$: $\widetilde{Y}_{\text{tar}} = Y_{\text{tar}} + \delta_Y$. Because the resulting perturbed trajectories $\widetilde{X}_{\text{tar}}$ and $\widetilde{Y}_{\text{tar}}$ need to be similar to the original trajectory, perturbation magnitudes are constrained to $\delta_X \in C_X$ and $\delta_Y \in C_Y$. 
Furthermore, to ensure dynamic feasibility of the perturbed trajectories, we follow Cao et al.~\cite{cao_advdo_2022} and perturb not the recorded positions $X_{\text{tar}}$ (which may result in trajectories that cannot be executed by real vehicles), but instead the control inputs $\bm{u} = \{a, \kappa\}$ underlying this trajectory. Here, control inputs include longitudinal acceleration $a$ and curvature $\kappa$.

\subsection{Dynamic model}
To allow us to perturb control actions, we must first be able to extract these control actions from a given trajectory $P = \{\bm{p}(t) \mid t\in \{T_0,\ldots,T_1\}\}$ (with positions $\bm{p}(t) = \{x(t), y(t)\}$), and transform the perturbed control actions into the perturbed trajectory. To this end, we use the dynamic model $\Phi$. This dynamic model allows us to advance the state $\bm{s} = \{x(t), y(t), \theta(t), v(t), \mathcal{D}(t)\}$ with heading angle $\theta$ and longitudinal velocity $v$:
\begin{equation}
    \bm{s}(t+1) = \Phi\left(\bm{s}(t), \bm{u}(t)\right) \,.
\end{equation}
Specifically, we get 
\begin{equation}
    \begin{aligned} \label{eq:Dynamic_model}
        v(t+1) & = v(t) + a(t) \Delta t \\
        \theta (t+1) & = \theta (t) + v(t) \kappa(t) \Delta t \\
        x(t+1) & = x(t) + v(t+1) \cos(\theta(t+1)) \Delta t \\
        y(t+1) & = y(t) + v(t+1) \sin(\theta(t+1)) \Delta t \,.
    \end{aligned}
\end{equation}
The dynamic model can also be inverted to give us the control actions, with $\bm{u}(t) = \Phi^{-1}(\bm{s}(t+1), \bm{s}(t))$, with
\begin{equation}
    \begin{aligned} \label{eq:Inverse_dynamic_model}
        a(t) & = \frac{v(t+1) - v(t)}{\Delta t} \\
        \kappa(t) & = \frac{\theta(t+1) - \theta(t)}{v(t)\Delta t} \,.
    \end{aligned}
\end{equation}

Many existing vehicle datasets only provide position data but not control input data. Consequently, to be able to use our dynamic model, we need to make the assumption that in the given unperturbed trajectory, intermediate states like velocities and heading angles are constant over the first two timestep (while this might not be true, there is no previous data to contradict this assumption). This allows us then to extract the initial state using equation \eqref{eq:Dynamic_model}, with $v_x(t) = \frac{x(t + 1) - x(t)}{\Delta t}$ and $v_y(t) = \frac{y(t + 1) - y(t)}{\Delta t}$:
\begin{equation}
    \begin{aligned} \label{eq:Inverse_assumption}
        v(T_0) & = \sqrt{v_x(T_0)^2 + v_y(T_0)^2} \\
        \theta(T_0) & = \text{atan2}\left(v_y(T_0), v_x(T_0)\right)
    \end{aligned}
\end{equation}
Additionally, in the definition of $\Phi^{-1}$ (equation \eqref{eq:Inverse_dynamic_model}), we cannot use $v(t+1)$ and $\theta(t+1)$, as (starting from $t = T_0$) we only have access to $v(t)$ and $\theta(t)$, requiring us to first calculate those (again using equation \eqref{eq:Dynamic_model}):
\begin{equation}
    \begin{aligned} \label{eq:Inverse_dynamic_model_actual} 
        v(t+1) & = \mathcal{D}_{t} \sqrt{v_x(t)^2 + v_y(t)^2} \\
        \theta(t+1) & = \text{atan2}\left(\mathcal{D}_{t}v_y(t), \mathcal{D}_{t}v_x(t)\right) \,. 
    \end{aligned}
\end{equation}
Here, $\mathcal{D}_{t}$ describes the current direction of travel (i.e., if the agent is reversing or driving forward), with
\begin{equation}
    \mathcal{D}_{t} = \begin{cases}
        \hphantom{-}\text{sgn}\left(v(t)\right)  & \vert  \text{atan2}\left(v_y(t), v_x(t)\right) - \theta(t) \vert \leq \frac{\pi}{2} \\
        -\text{sgn}\left(v(t)\right) & \text{otherwise} \end{cases} \,.
\end{equation}
This ensures that reversing the direction of travel is correctly extracted as a (sign) change in velocity brought about by realistic accelerations or decelerations and not as a turn that would require an impossibly high curvature.

\subsection{Generating dynamically feasible attacks}
\begin{figure*}    
    \centering
    \includegraphics[width = \textwidth]{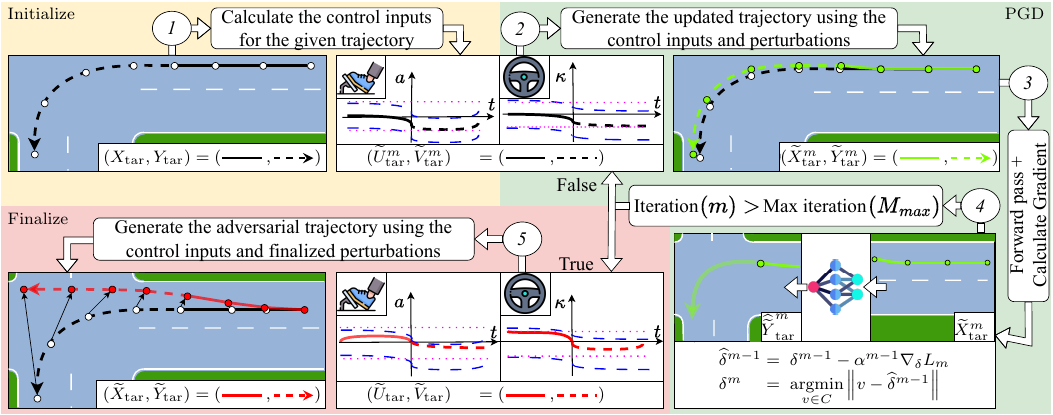}
    \caption{Generating adversarial trajectories with constraining future states using control actions. 1) During initialization, we extract the control actions ($U_{\text{tar}} = \widetilde{U}^0_{\text{tar}}$, $V_{\text{tar}} = \widetilde{V}_{\text{tar}}^0$) from the unperturbed trajectory parts $X_{\text{tar}}$ and $Y_{\text{tar}}$ respectively. 2)
    Afterwards, we iteratively perturb those control actions, where, starting with the given perturbed control actions $\widetilde{U}_{\text{tar}}^m$ and $\widetilde{V}_{\text{tar}}^m$, we get the corresponding perturbed trajectories $\widetilde{X}_{\text{tar}}^m$ and $\widetilde{Y}_{\text{tar}}^m$. 3) Based thereon -- together with the prediction $\widehat{\widetilde{Y}}^m_{\text{tar}}$ from updated states $\widetilde{X}_{\text{tar}}^m$ -- we can calculate the adversarial loss $L_m$ and its derivative, with which the perturbation can be updated. The constraints $C$ in this update are represented by the purple dotted line for the absolute control limits and the blue dashed line for the relative bounds. 4) This procedure is repeated for $M_{\max}$ iterations. 5) In the end, the final perturbed trajectories are extracted.}
    \label{Fig:Dynamics_explained_new}
\end{figure*}
With the above dynamic model, we can generate dynamically feasible attacks in an iterative procedure (Figure~\ref{Fig:Dynamics_explained_new}).
\begin{enumerate}
    \item In the first iteration ($m=0$), we extract the control actions $U_{\text{tar}} = \{\bm{u}_{\text{tar}}(t) \mid t \in \{-H+1,\ldots, -1\}$ and $V_{\text{tar}} = \{\bm{u}_{\text{tar}}(t) \mid t \in \{0, \ldots, T-1\}$ from the unperturbed trajectories $X_{\text{tar}}$ and $Y_{\text{tar}}$ respectively. After initializing the added states (equation~\eqref{eq:Inverse_assumption}), we can use alternately the inverse model $\Phi^{-1}$ (equations~\eqref{eq:Inverse_dynamic_model} and~\eqref{eq:Inverse_dynamic_model_actual}) to get the current control actions, before using the forward model $\Phi$ (equation~\eqref{eq:Dynamic_model}) to the get the next values of the added states (Figure~\ref{Fig:Dynamics_explained_new}.1). 
    We also initialize our control action perturbations $\delta_{U_{\text{tar}}}^0 = \bm{0}$ and $\delta_{V_{\text{tar}}}^0 = \bm{0}$.

    \item We get the perturbed control states $\widetilde{U}^m_{\text{tar}} = U_{\text{tar}} + \delta^m_{U_{\text{tar}}}$ and $\widetilde{V}^m_{\text{tar}} = V_{\text{tar}} + \delta^m_{U_{Y_{\text{tar}}}}$. These can then be used to compute the corresponding perturbed trajectories $\widetilde{X}^m_{\text{tar}}$ and $\widetilde{Y}^m_{\text{tar}}$, using the full initial state $\bm{s}_{\text{tar}}(-H+1)$ (equation~\eqref{eq:Inverse_assumption}) and the given control inputs together with the dynamic model $\Phi$ (see Figure~\ref{Fig:Dynamics_explained_new}.2). 

    \item We use the prediction model to generate the corresponding predictions $\widehat{\widetilde{Y}}^m_{\text{tar}}$ based on the perturbed input. Given an adversarial loss function $L^m = \mathcal{L}\left(X,Y, \widetilde{X}^m_{\text{tar}}, \widetilde{Y}^m_{\text{tar}}, \widehat{\widetilde{Y}}^m_{\text{tar}} \right)$, we can get the gradient of the loss in regard to the perturbation $\delta^m_{U_{\text{tar}}}$ and $\delta^m_{V_{\text{tar}}}$ using differentiable programming (see Appendix~\ref{app:derivatives_new} for details). With this gradient, we can use projected gradient descent to update the perturbations $\delta^m_{U_{\text{tar}}}$ and $\delta^m_{V_{\text{tar}}}$ (Figure~\ref{Fig:Dynamics_explained_new}.3).
    
    In regard to the constraints $C_{U}$ and $C_{V}$ (equation~\eqref{eq:PGD_2}), we choose the following values to restrict the perturbations of  acceleration $a(t)$ and  curvature $\kappa(t)$ 
    \begin{equation}
        \begin{aligned} \label{eq:control_constraints}
            \delta_a(t) & \geq \max \{\SI{-2}{ms^{-2}}, a_{\min} - a(t)\} \\
            \delta_a(t) & \leq  \min \{\SI{2}{ms^{-2}}, a_{\max} - a(t)\} \\
            \delta_\kappa(t) & \geq  \max \{\SI{-0.05}{m^{-1}}, \SI{-0.2}{m^{-1}} - \kappa(t)\} \\
            \delta_\kappa(t) & \leq  \min \{\SI{0.05}{m^{-1}}, \SI{0.2}{m^{-1}} - \kappa(t)\}
        \end{aligned}
    \end{equation}
    This approach combines both absolute bounds (i.e., bounds on the resulting perturbed control inputs $\widetilde{a}(t)$ and $\widetilde{\kappa}(t)$) as well as relative bounds (directly on perturbations $\delta_a(t)$ and $\delta_{\kappa}(t)$, taken from~\cite{cao_advdo_2022}). For the absolute bounds $a_{\min}$ and $a_{\max}$, we use the lowest and highest acceleration values observed in the dataset.

    \item If $m>M_{\max}$, we complete the attack, having arrived at our final perturbed trajectories $\widetilde{X} = \widetilde{X}^{M_{\max}}$ and $\widetilde{X} = \widetilde{X}^{M_{\max}}$ (Figure~\ref{Fig:Dynamics_explained_new}.5).
\end{enumerate}

\subsection{Preserving Tactical Behavior} \label{ch:regularization}
As discussed in Section~\ref{ch:introduction}, in order for the adversarial attack to be effective, perturbed trajectories need to provide for the same tactical behavior as the ground truth, i.e., the should be close to  the original observations. However, our choice of perturbing control actions instead of positions makes it impossible to implement this via the constraints $C_{U}$ and $C_{V}$ (equation~\eqref{eq:control_constraints}).
Therefore, we chose to instead implement such position constraints by adding a regularizing term $l$ to the adversarial loss $\mathcal{L}$. Here, a cost function $l$ constraining a whole trajectory combines costs $l_p$ that constrain each individual perturbed position $\widetilde{\bm{p}}(t)$ of the original trajectory $P = \{{\bm{p}}(t) \mid t \in \{T_0,\ldots,T_1\}\}$.

\begin{figure*}
    \centering
    \includegraphics[width=1\linewidth]{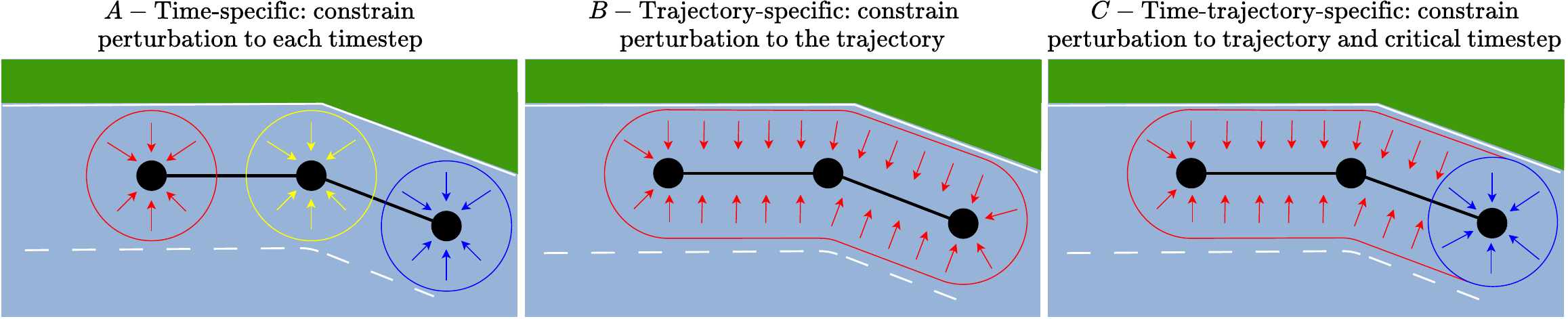}
    \caption{Three regularization approaches for preserving tactical behavior. The black line represents the ground truth trajectory, with the colored arrows showing the direction in which the regularization penalties are applied. In A), the different colors represent that the different penalties are applied to different timesteps.}
    \label{Fig:regularization}
\end{figure*}

\subsubsection{Regularizing individual points in the perturbed trajectory}
Inspired by negative log-barrier constraint encoding used in interior point methods~\cite{boyd2004convex}, we use the following regularization term:
\begin{equation} \label{eq:regularization_point}
    l_{p,\text{Method}}\left(\widetilde{\bm{p}}(t), P\right) = - \ln\left(d_{\max} - d_{\text{Method}}(\widetilde{\bm{p}}(t), P)\right)
\end{equation}
Importantly, $l_{p,\text{Method}}\left(\widetilde{\bm{p}}(t),P\right)$ grows to infinity as $d_{\text{Method}}(\widetilde{\bm{p}}(t), P)$ approaches $d_{\max}$. Therefore, with sufficiently small steps, the constraint $d_{\text{Method}}(\widetilde{\bm{p}}(t), P) \leq d_{\max}$ will never be violated. Here, Method denotes one of the two options for the choice of the distance function:
\begin{itemize}
    \item \textbf{Time-specific}: The perturbed position is bounded by the corresponding unperturbed position:
    \begin{equation} \label{eq:regularization_point_time}
        d_{\text{Time}}(\widetilde{\bm{p}}(t), P) = \Vert \widetilde{\bm{p}}(t) - \bm{p}(t) \Vert
    \end{equation}
    \item \textbf{Trajectory-specific}: The perturbed trajectory is bounded in its shortest distance to the whole ground truth trajectory; only lateral perturbations are penalized.

    \begin{equation} \label{eq:regularization_point_traj}
        d_{\text{Traj}}(\widetilde{\bm{p}}(t), P) = \min_{t'} d_{\text{proj}}(\widetilde{\bm{p}}(t), \bm{p}(t'), \bm{p}(t' + 1)) 
    \end{equation}
    Here $d_{\text{proj}}$ is the closest distance of point onto a line segment spanned by two other points:
    \begin{equation*}
        d_{\text{proj}}(\bm{a}, \bm{b}, \bm{c}) = 
        \begin{cases} 
        \Vert \bm{a} - \bm{c} \Vert  & r(\bm{a}, \bm{b}, \bm{c}) \leq 0 \\
        d_{\perp}(\bm{a}, \bm{b}, \bm{c}) & 0 < r(\bm{a}, \bm{b}, \bm{c}) < 1 \\
        \Vert \bm{a} - \bm{b} \Vert  & \text{otherwise}.
        \end{cases}
    \end{equation*}
    with 
    \begin{equation*}
    \begin{aligned}
        d_{\perp}(\bm{a}, \bm{b}, \bm{c}) & = \frac{\Vert (\bm{a} - \bm{c}) \times (\bm{b} - \bm{c})\Vert}{\Vert (\bm{b} - \bm{c})\Vert}  \\
        r(\bm{a}, \bm{b}, \bm{c}) & = \frac{(\bm{a} - \bm{c}) \cdot (\bm{b} - \bm{c})}{\Vert (\bm{b} - \bm{c})\Vert^2}
    \end{aligned}
    \end{equation*}
    
\end{itemize}

\subsubsection{Regularizing perturbed trajectories} \label{sec:position_constraints}
Based on different combinations of the pointwise constraints discussed above, we propose three different approaches for regularizing full trajectories $P$ (Figure~\ref{Fig:regularization}):
\begin{itemize}
    \item \textbf{Time-specific}: only use the time-specific constraints (equation~\eqref{eq:regularization_point_time}):
    \begin{equation} \label{eq:regularization_time}
        l_{\text{Time}}(\widetilde{P}, P) = \frac{1}{\vert \widetilde{P}\vert} \sum\limits_{\widetilde{\bm{p}}(t) \in \widetilde{P}} l_{p,\text{Time}}(\widetilde{\bm{p}}(t), P)
    \end{equation}
    \item \textbf{Trajectory-specific}: only use the trajectory-specific constraints (equation~\eqref{eq:regularization_point_traj}):
    \begin{equation}\label{eq:regularization_traj}
        l_{\text{Traj}}(\widetilde{P}, P) = \frac{1}{\vert \widetilde{P}\vert} \sum\limits_{\widetilde{\bm{p}}(t) \in \widetilde{P}} l_{p,\text{Traj}}(\widetilde{\bm{p}}(t), P) 
    \end{equation}
    \item \textbf{Time-trajectory-specific}: While the trajectory-specific approach allows for more flexibility in longitudinal deviations, thereby enabling the exploration of a broader range of realistic adversarial scenarios, this also poses a risk. Namely, perturbations may lead to substantial deviations from the ground truth states at critical moments, resulting in unrealistic or unsafe behaviors not aligned with the natural flow of the trajectory. For example, when making a left turn (Figure~\ref{Fig:Introduction_image}A), it is crucial that, at prediction time $t=0$, the target agent can still feasibly complete the turn without risking a collision with the ego agent.
    Therefore we introduce a combined regularization function
    \begin{equation}\label{eq:barrier_function_advanced_new}
    \begin{split}
        l_{\text{Time-traj}}(\widetilde{P}, P) = l_{\text{Traj}}(\widetilde{P}, P) + l_{p,\text{Time}}(\widetilde{\bm{p}}(T_1), P) \,.
    \end{split}
    \end{equation}
\end{itemize}

\subsection{Adversarial attack objectives} \label{ch:new_attack_strategies_comp}
For our new attack generation approach, we implemented the attack objectives common in the literature: ADE/FDE and collision rate (Section~\ref{ch:attack_functions_PGD}). Furthermore, we also propose a novel attack objective: \emph{false negative collision}. In contrast to the traditional (\emph{false positive}) collision attack (which tries to deceive the AV  into predicting a collision), our novel attack tries to conceal an intended collision from the AV. Compared to the false positive collision loss in equation~\eqref{eq:collision_simple}, the attack objective here is
\begin{equation}\label{eq:collision_hide}
\begin{aligned}
    l_{\text{FNC}}(Y_{\text{ego}},{\widetilde{Y}}_{\text{tar}}, \widehat{\widetilde{Y}}_{\text{tar}}) & =  \min_{t \in \{1, \ldots, T\}} \left\Vert \widetilde{\bm{y}}_{\text{tar}}^k(t) - \bm{y}_{\text{ego}}(t) \right\Vert \\ & \hphantom{=} + \frac{1}{T} \sum_{t=1}^{T} \left\Vert \frac{1}{K} \sum_{k=1}^{K} \widehat{\widetilde{\bm{y}}}_{\text{tar}}^{k}(t) - \widehat{\bm{y}}_{\text{tar}}^k(t) \right\Vert
\end{aligned}
\end{equation}
The first term here is designed to create a perturbed future that seeks a collision, while the second term tries to create a perturbation for which the attacked model's predictions are similar to the prediction resulting from the ground truth (i.e., the prediction model should not be able to differentiate between them). The approach of deceiving the prediction model into making specific predictions was originally proposed by Tan et al. \cite{tan_targeted_2022}.
The value of this novel attack is that it presents a far more dangerous scenario for an \acrshort{av} in actual traffic  than the traditional collision attack. Namely, in the false positive collision case, the reaction of the \acrshort{av} would be an unnecessary braking maneuver, while a successful false negative collision attack can be far more severe, especially if the vehicles are close enough that there is not enough time left for a safe evasion upon the \acrshort{av} recognizing the attack.

\section{Experimental Setting} \label{ch:data}
We design a number of experiments to test different attack profiles, using an evaluation Framework designed by Schumann \emph{et al.}~\cite{schumann2023benchmarking}.\footnote{The code for running these experiments can be found online: \url{https://github.com/jhagenus/General-Framework-update-adversarial-Jeroen/blob/main/README.md}}
\subsection{Trajectory prediction model} As a case study, we applied our attack generation approach to one of the most popular trajectory prediction models: \emph{Trajectron++}~\cite{salzmann_trajectron_2020}. \emph{Trajectron++}, predicting not positions directly, but instead control states of a unicycle model, has the advantage of always producing dynamically feasible predictions, which makes it a particularly suitable model to test adversarial attacks.
We set the model to generate $K=100$ predictions for each model input, i.e., in each scene.

\subsection{Dataset}
As a source of unperturbed trajectories, we chose 
the L-GAP~\cite{zgonnikov_should_2024} dataset. This urban driving dataset obtained in a driving simulator focuses on the intersection scenario. Sixteen human drivers (yellow vehicle in Figure~\ref{Fig:Introduction_image}A) were driving through an urban environment, executing a large number of left turns either in front of or behind an oncoming AV (blue vehicle), which was programmed to always go straight with constant velocities. We selected the left-turning vehicle as the target agent for which we perturb the trajectory. The oncoming vehicle approaching the intersection 
was designated as the ego vehicle, which would make a prediction about the behavior of the target agent. Because of noisy measurement of vehicle coordinates in the original dataset, we used the Savitzky-Golay filter to smooth the trajectories.

Unlike most real-life datasets that aim to cover diverse traffic scenarios, the L-GAP dataset focuses on a specific scenario, and provides a large number of trajectories for that scenario (we used 1024 left-turn trajectories produced by 16 human drivers). However, this introduces a data bias that can significantly affect model training. Because the target agent either makes a left turn or waits at the crossing, a prediction model trained on this dataset alone would have difficulties properly handling different maneuvers which might arise from adversarial attacks. To address this issue, we trained the trajectory prediction model simultaneously on both L-GAP and NuScenes~\cite{caesar_nuscenes_2020}, with the latter being larger and including a far more varied range of scenarios. However, adversarial attacks were only generated using trajectories from L-GAP to facilitate interpretation of evaluation results.

In this work, we used an observation period $H=12$ time steps, and a prediction horizon $T=12$ time steps, with a time interval of $\Delta t = 0.1s$. These parameters were chosen based on the sampling length of the dataset. To make the prediction scenario as relevant as possible, we evaluated predictions at a point in time when the time-to-arrival of the ego agent to the intersection is specifically so large that it could barely come to a stop before the intersection if a collision is predicted. Consequently, a wrong prediction would make the ago agent vulnerable to a potentially unavoidable collision, therefore increasing the importance of a correct prediction.

\subsection{Comparison to established attack approaches} 
We compared our method of perturbing control actions with constraints on future states to two alternative approaches commonly used in the literature~\cite{hagenus2024survey}. In the first one, one will simply perturb positions, without any constraints regarding dynamic feasibility. The second approach is the \textit{search attack}proposed by Zhang et al. \cite{zhang2022adversarial}, which also perturbs spatial positions but limits the perturbation using data statistics. Unlike our approach, both of these approaches impose constraints only on past states.

For each of the three attack strategies (positions; positions with data statistics~\cite{zhang2022adversarial}; actions with future state constraints (ours)), we tested all possible combinations of four attack objectives (ADE; FDE; false positive collision; false negative collision) and two constraint types (time-specific; time-trajectory-specific). This resulted in eight adversarial attack loss functions for each attack approach.

Our approach includes perturbing not only observed (past) but also future states; for perturbing the latter, we used the trajectory-specific displacement constraints (equation~\eqref{eq:regularization_traj}). The reason for using only this constraint type is that the future perturbed trajectories $\widetilde{Y}_{\text{tar}}$ should mostly preserve the actually chosen maneuver, for which this looser constraint is enough.

Finally, to highlight the added value of perturbing control actions with trajectory-specific constraints on future states, we compared our approach to its ablated version attacking control actions without constraining future states.

\subsection{Adversarial attack parameters} 
For all loss functions, we set the maximum displacement to $d_{\text{Max}} = \SI{0.9}{m}$. This choice is based on a lane width of $\SI{3.5}{m}$ and vehicle width of $\SI{1.7}{m}$, so that even after perturbations, a vehicle should not leave its lane. 

To generate attacks with projected gradient descent, we used an initial learning rate $\alpha^0 = 0.01$ (equation~\eqref{eq:PGD_1}), with a exponential decay of $\gamma = 0.99$ (i.e., $\alpha^m = \gamma \alpha^{m-1}$) for $M_{\max} = 100$ iterations
Crucially, $\alpha$ can be adjusted if the perturbed observed and future states fall outside the tactical behavior constraint region (i.e., in equation~\eqref{eq:regularization_point}, we see $d(\widetilde{\bm{p}}(t), P) > d_{\max}$). In such cases, we repeatedly halved $\alpha$ until this condition was met. This learning rate scaling was applied only within each iteration, so that the following iteration would still use the same learning rate as if no adjustment was necessary on the previous iteration.

\subsection{Performance measures} 
We employed eight performance measures to evaluate our proposed attack approach as well as established methods. We used measure values \emph{averaged} across all the samples in the dataset.

\subsubsection*{Attack efficiency}
We adopted three measures that are widely used in the literature~\cite{hagenus2024survey}:
\begin{itemize}
    \item Average displacement error:
    \begin{equation}
    \text{ADE} = \frac{1}{K T} \sum_{k=1}^{K} \sum_{t=1}^{T}  \left\Vert \widehat{\widetilde{\bm{y}}}_{\text{tar}}^k(t) - \bm{y}_{\text{tar}}(t) \right\Vert
    \end{equation}

    \item Final displacement error:
    \begin{equation}
    \text{FDE} = \frac{1}{K} \sum_{k=1}^{K}\left\Vert \widehat{\widetilde{\bm{y}}}_{\text{tar}}^k(T) - \bm{y}_{\text{tar}}(T) \right\Vert
    \end{equation}
    
    \item Collision rate:
    \begin{equation}
    \text{CR}_{\text{pred}} = \frac{1}{K} \sum_{k=1}^K \underset{t \in \{1, \ldots, T\}}{\max} f_{\text{coll}}\left(\widehat{\widetilde{y}}_{\text{tar}}^k(t),y_{\text{ego}}(t)\right)
    \end{equation} with
    \begin{equation}
    f_{\text{coll}}(\bm{a}, \bm{b}) = \begin{cases}
        1 & \text{BB}\left(\bm{a}\right) \cap \text{BB}\left(\bm{b}\right) \neq \emptyset \\
        0 & \text{otherwise}
    \end{cases} \, ,
    \end{equation}
    where $\text{BB}(\cdot)$ represents the bounding box of a vehicle.
\end{itemize}

To accompany our novel \emph{false negative collision} attack, we propose a new measure
\begin{itemize}
    \item Rate of collisions along the perturbed path:
    \begin{equation}
    \text{CR}_{\text{FNC}} = \underset{t \in \{1, \ldots, T\}}{\max} f_{\text{coll}}\left({\widetilde{y}}_{\text{tar}}(t),y_{\text{ego}}(t)\right) \,.
    \end{equation}
\end{itemize}

\subsubsection*{Perturbation magnitude}
We used two measures to measure the extent of the produced perturbations, with larger values corresponding to more noticeable attacks:
\begin{itemize}
    \item Maximum displacement: 
    \begin{equation}
        D_{\text{max}} = \max_{\{t \in {-H+1, \ldots, 0}\}} \Vert \widetilde{\bm{x}}_{\text{tar}}(t) - \bm{x}_{\text{tar}}(t) \Vert
    \end{equation}
    \item Mean displacement: 
    \begin{equation}
        D_{\text{mean}} = \frac{1}{H} \sum\limits_{t=-H+1}^{0} \Vert \widetilde{\bm{x}}_{\text{tar}}(t) - \bm{x}_{\text{tar}}(t) \Vert
    \end{equation}
\end{itemize}

\subsubsection*{Dynamic feasibility}
We used two measures to evaluate the dynamic feasibility of the perturbed past observations $\widetilde{X}_{\text{tar}}$ using the perturbed control inputs in $\widetilde{U}_{\text{tar}}$:
\begin{itemize}
    \item Acceleration magnitude:
    \begin{equation}
    a = \frac{1}{H-1} \sum_{t=-H+1}^{-1} |\widetilde{a}(t)|
    \end{equation}

    \item Curvature magnitude: 
    \begin{equation}
    \kappa = \frac{1}{H-1} \sum_{t=-H+1}^{-1} |\widetilde{\kappa}(t)|
    \end{equation}
\end{itemize}

\section{Results}
We evaluated attack efficiency, the magnitude of generated perturbations, and dynamic feasibility of adversarial trajectories for three attack approaches: position-based attack (Table~\ref{tab:Performance_attacks_position}), search attack perturbing positions using data statistics-based constraints (Table~\ref{tab:Performance_attacks_search}), and our approach of perturbing control action with constraints on future states (Table~\ref{tab:Performance_attacks_our}).

\begin{table*}
\centering
\caption{Evaluation of perturbing observed positions (without using constraints on future behavior).}
\label{tab:Performance_attacks_position}
\begin{tabular}{ll||cccc|cc|cc}
\toprule
 & Position constraints & \multicolumn{4}{c|}{Attack efficiency} & \multicolumn{2}{c|}{Perturbation magnitude} & \multicolumn{2}{c}{Dynamic feasibility} \\ 
Attack loss & (observed) & ADE [$m$] & FDE [$m$] & $\text{CR}_{\text{pred}}$ & $\text{CR}_{\text{FNC}}$ & $D_{\text{max}}$ [$m$] & $D_{\text{mean}}$ [$m$] & $a$ [$ms^{-2}$] & $\kappa$ [$m{-1}$] \\
\midrule
Unperturbed &   -  & 0.115 & 0.283 & 0.015 & - & - & - & 0.510 & 0.006 \\
\midrule
ADE & Time & 4.045 & 3.447 & 0.007 & - & 0.770 & 0.195 & 53.407 & 20.860 \\
   & Time-traj & 3.447 & 6.203 & 0.011 & -  & 0.681 & 0.168 & 47.066 & 25.222 \\
\midrule
FDE   & Time & 3.983 & 7.158 & 0.008 & - & 0.766 & 0.186 & 51.868 & 23.368 \\
   & Time-traj &4.035 & 7.262 & 0.008 & - & 0.846 & 0.197 & 56.280 & 22.338 \\
\midrule
False Positive  & Time & 1.371 & 2.495 & 0.020 & - & 0.467 & 0.119 & 36.695 & 27.427 \\
Collision & Time-traj & 1.389 & 2.530 & 0.020 & - & 0.495 & 0.125 & 39.004 & 29.138 \\
\midrule
False Negative   & Time & 0.562 & 1.020 & 0.019 & 0.736 & 0.423 & 0.116 & 36.427 & 25.177 \\
Collision   & Time-traj & 0.598 & 1.085 & 0.020 & 0.741 & 0.467 & 0.126 & 40.322 & 25.739 \\
\bottomrule
\end{tabular}
\end{table*}

\begin{table*}
\centering
\caption{Evaluation of the position-based ``search'' attack~\cite{zhang2022adversarial} (without using constraints on future behavior).}
\label{tab:Performance_attacks_search}
\begin{tabular}{ll||cccc|cc|cc}
\toprule
 & Position constraints & \multicolumn{4}{c|}{Attack efficiency} & \multicolumn{2}{c|}{Perturbation magnitude} & \multicolumn{2}{c}{Dynamic feasibility} \\ 
Attack loss & (observed) & ADE [$m$] & FDE [$m$] & $\text{CR}_{\text{pred}}$ & $\text{CR}_{\text{FNC}}$ & $D_{\text{max}}$ [$m$] & $D_{\text{mean}}$ [$m$] & $a$ [$ms^{-2}$] & $\kappa$ [$m{-1}$] \\
\midrule
Unperturbed &  -  & 0.115 & 0.283 & 0.015 & - & - & - & 0.510 & 0.006 \\
\midrule
ADE  &  Time & 3.073 & 5.521 & 0.007 & - & 0.714 & 0.177 & 49.392 & 20.100 \\
  &  Time-traj & 2.028 & 3.668 & 0.010 & - & 0.539 & 0.131 & 38.738 & 27.571 \\
\midrule
FDE  & Time & 2.283 & 4.124 & 0.007 & - & 0.594 & 0.148 & 43.665 & 23.848 \\
 & Time-traj & 2.306 & 4.165 & 0.007 & - & 0.623 & 0.152 & 45.145 & 23.805 \\
\midrule
False Positive  & Time & 1.792 & 3.241 & 0.014 & - & 0.514 & 0.123 & 36.494 & 27.938 \\
Collision & Time-traj & 1.819 & 3.290 & 0.015 & - & 0.548 & 0.130 & 38.991 & 25.836 \\
\midrule
False Negative  & Time & 0.878 & 1.586 & 0.020 & 0.709 & 0.452 & 0.122 & 38.110 & 24.988 \\
Collision  & Time-traj & 0.933 & 1.685 & 0.019 & 0.709 & 0.481 & 0.130 & 40.906 & 24.469 \\
\bottomrule
\end{tabular}
\end{table*}

\begin{table*}
\centering
\caption{Evaluation of perturbing control actions with trajectory-specific constraints on future states (Ours). \\ Rows with ``-'' under "Future" display results for perturbing control actions without constraints on future states. }
\label{tab:Performance_attacks_our}
\begin{tabular}{lll||cccc|cc|cc}
\toprule
& \multicolumn{2}{c||}{Position constraints} & \multicolumn{4}{c|}{Attack efficiency} & \multicolumn{2}{c|}{Perturbation magnitude} & \multicolumn{2}{c}{Dynamic feasibility} \\  
Attack loss  & Observed & Future & ADE [$m$] & FDE [$m$] & $\text{CR}_{\text{pred}}$ & $\text{CR}_{\text{FNC}}$ & $D_{\text{max}}$ [$m$] & $D_{\text{mean}}$ [$m$] & $a$ [$ms^{-2}$] & $\kappa$ [$m{-1}$] \\
\midrule
Unperturbed &  -  &  -  & 0.115 & 0.283 & 0.015 & - & - & - & 0.510 & 0.006 \\
\midrule
ADE  & Time &  -  & 1.030 & 1.604 & 0.014 & - & 0.841 & 0.525 & 1.658 & 0.006 \\
  & Time & Trajectory& 0.250 & 0.470 & 0.021 & - & 0.150 & 0.054 & 0.901 & 0.005 \\
  & Time-traj &  -  & 0.738 & 1.201 & 0.017 & - & 0.525 & 0.153 & 1.402 & 0.005 \\
  & Time-traj & Trajectory& 0.222 & 0.435 & 0.021 & - & 0.113 & 0.038 & 0.878 & 0.005 \\
\midrule
FDE & Time &  -  & 1.059 & 1.651 & 0.015 & - & 0.856 & 0.267 & 1.792 & 0.007 \\
 & Time & Trajectory& 0.581 & 0.943 & 0.021 & - & 0.459 & 0.159 & 1.112 & 0.006 \\
& Time-traj &  -  & 0.841 & 1.360 & 0.017 & - & 0.598 & 0.161 & 1.644 & 0.006 \\
 & Time-traj & Trajectory& 0.503 & 0.845 & 0.021 & - & 0.349 & 0.108 & 1.113 & 0.005 \\
\midrule
False Positive & Time &  -  & 0.446 & 0.715 & 0.021 & - & 0.360 & 0.127 & 1.065 & 0.006 \\
Collision  & Time & Trajectory& 0.295 & 0.504 & 0.020 & - & 0.214 & 0.077 & 0.915 & 0.006 \\
 & Time-traj &  -  & 0.332 & 0.570 & 0.021 & - & 0.217 & 0.067 & 1.005 & 0.005 \\
 & Time-traj & Trajectory& 0.255 & 0.459 & 0.020 & - & 0.152 & 0.049 & 0.905 & 0.006 \\
\midrule
False Negative  & Time &  -  & 0.123 & 0.222 & 0.018 &  0.723 & 0.074 & 0.028 & 0.845 & 0.006 \\
Collision  & Time-traj &  -  & 0.105 & 0.198 & 0.018 & 0.723 & 0.052 & 0.021 & 0.855 & 0.006 \\
\bottomrule
\end{tabular}
\end{table*}

\subsection{Position-based attacks produce dynamically impossible trajectories}

We found that both the position attack and the search attack yielded significantly higher \acrshort{ade} and \acrshort{fde} values compared to our approach. For example, ADE attack with time-specific position constraints resulted in the \acrshort{ade} value of $\SI{4.045}{m}$ for the position-based attack, $\SI{3.073}{m}$ for the search attack, and only $\SI{1.030}{m}$/$\SI{0.250}{m}$ (without/with constraints on future positions) for an attack on control actions, compared to $\SI{0.115}{m}$ that the model achieves on unperturbed inputs. 

However, the high efficiency of position-based attacks was achieved by producing trajectories that are dynamically impossible. For both established methods (Tables~\ref{tab:Performance_attacks_position}, \ref{tab:Performance_attacks_search}), the \emph{average} acceleration magnitude $a$ exceeds $\SI{30}{ms^{-2}}\approx 3g$ for all attack losses with average trajectory curvature of at least $\SI{20}{m^{-1}}$ (requiring a turn radius lower than $\SI{0.05}{m}$).
Trajectories with such extreme acceleration and curvature values cannot be reproduced in real vehicles(see Supplementary Information Figures~SI\ref{Fig:ADE_Position}~and~SI\ref{Fig:ADE_Search} for examples) and hence do not provide a valid means to assess adversarial robustness of trajectory prediction models. 

\subsection{Constraining future states generates efficient and realistic control action-based attacks}
\label{sec:Future_constraint_results}
The above results demonstrate that the established attacks achieve high efficiency but are by far not realistic. In contrast, our approach of perturbing control actions (instead of positions directly) produces actions that are similar in magnitude to the unperturbed data (Table~\ref{tab:Performance_attacks_our}), and therefore can be implemented in a real vehicle. Although the impact of the attacks generated by this approach is much lower than for established approaches (as indicated by lower attack efficiency measures), these attacks still lead to substantially decreased accuracy of the attacked model. For instance, the ADE and FDE values achieved by attacks targeting these measures are still substantially higher than for the predictions based on unperturbed trajectories. Importantly, false positive collision attacks for our approach generate similar (compared to position-only constraints) or higher (compared to the search attack) predicted collision rates.

Further investigating the approach of perturbing control actions, we analyzed the effect of using future position constraints on attack performance (Table~\ref{tab:Performance_attacks_our}).
We found that imposing constraints on future positions (in addition to constraining  observed positions) results in perturbed trajectories which are far more similar to the ground truth. For instance, we see an decrease in acceleration magnitude of around 40\% for \acrshort{ade} and \acrshort{fde} attacks, and 10\% for a false positive collision attack. 

Similar effects were observed for the displacement measures. For past-only constraints, trajectories attacking \acrshort{ade} and \acrshort{fde} consistently approached the maximum allowed displacement of $\SI{0.9}{m}$ (with values of $D_{\text{max}} = \SI{0.841}{m}$). For attacks generated by constraining future states however, the displacement values are consistently and substantially lower, resulting in adversarial trajectories that closely align with original, unperturbed ones. This is likely because adding the constraints on future perturbed trajectories impose further limits on the latter timesteps of the observed trajectory (in Supplementary Information, compare Figures~SI\ref{Fig:ADE_control_action}~and~SI\ref{Fig:feasible_ade_control_action} or Figures~SI\ref{Fig:FDE_control_action}~and~SI\ref{Fig:feasible_fde_control_action} for illustrative examples).
The direct consequence of these smaller perturbation magnitudes however is the reduced efficiency, resulting in lower \acrshort{ade} and \acrshort{fde} values across all attack objectives. 

These results highlight a trade-off between attack efficiency (how likely an attack is to deceive a prediction model) and obscurity (how difficult it is to detect the attack). However, surprisingly, we found that attacking the collision rate does not conform to this trade-off.  
Under the false positive collision attack, constraining future states improved the dynamic feasibility and reduced attack magnitude but resulted in similar $\text{CR}_{\text{pred}}$ values. Even more surprisingly, for  \acrshort{ade} and \acrshort{fde} attacks, collision rates actually increased when constraining future states. The reason for this is that in the left turn scenario, our attack has to deceive the model into predicting a turn that is either sharper or wider than the one pursued originally. With curvature values being consistent with the dataset, it is possible that both options are regularly chosen. For wider turns, those can then be wide enough to avoid the oncoming vehicle, with the target agent staying in its own lane. However, if future perturbations and therefore prediction errors are constrained, achieving this is less likely, which makes predicted collision more likely (see again Supplementary Information Figures~SI\ref{Fig:ADE_control_action}~and~SI\ref{Fig:feasible_ade_control_action} for an example).  

\subsection{Attacking the false negative collision rate is highly efficient}
Our novel false negative collision attack was remarkably successfully in deceiving the model into believing that a perturbed trajectory is not on a collision course, while it actually was. Across all three approaches (Tables~\ref{tab:Performance_attacks_position}, \ref{tab:Performance_attacks_search}, \ref{tab:Performance_attacks_our}), at least 70\% of generated trajectories were on a collision course while the model believed they were not. Moreover, for the control action-based perturbations (Table~\ref{tab:Performance_attacks_our}), the adversarial attacks effectively deceived the model into making predictions $\widehat{\widetilde{Y}}$ that roughly resemble the predictions based on the ground truth $\widehat{Y}$. Here, trajectories generated using the false negative collision loss were indistinguishable from the ground truth, with \acrshort{ade} and \acrshort{fde} values similar or lower than on unperturbed inputs (see also Supplementary Information Figure~SI\ref{Fig:Hide_collision_control_action}). Importantly, highlighting the danger of this attack, the model was deceived only \SI{1.2}{s} before a potential collision, leaving little if not no time to correct that assessment.

\subsection{Regularization approach affects performance of control action-based but not position-based attacks} 
For the control action-based attacks (Table~\ref{tab:Performance_attacks_our}), 
using the time-trajectory specific constraints (equation~\eqref{eq:barrier_function_advanced_new}) generally resulted in lower attack efficiency measures compared to the time-specific approach, while sometimes slightly increasing the collision rates, corresponding to a drop in perturbation magnitude. For example, under the \acrshort{ade} attack with constrained future trajectories (Table~\ref{tab:Performance_attacks_our}), the \acrshort{ade} measure drops from $\SI{0.250}{m}$ to $\SI{0.222}{m}$ and \acrshort{fde} drops from $\SI{0.470}{m}$ to $\SI{0.435}{m}$, while maximum perturbation decreases from $\SI{0.150}{m}$to $\SI{0.113}{m}$ and average acceleration magnitude from $\SI{0.901}{ms^{-2}}$ to $\SI{0.878}{ms^{-2}}$. On first glance, this might seem somewhat counterintuitive, as the use of the full trajectory-based constraint should allow greater freedom. However, this is not unexpected, as we in this case we effectively doubled the constraint cost imposed on the last time step (equation~\eqref{eq:barrier_function_advanced_new}), which, due to the constraints on the control actions for dynamic feasibility also constrains the previous time steps, preventing them from exploiting their increased freedom.  

This assumption is supported by the fact that we did not observe similar trends in the position-based attacks (Tables~\ref{tab:Performance_attacks_position}~and~\ref{tab:Performance_attacks_search}), where this trend is not seen, and both constraint types could produce larger or smaller perturbations. Here, the attack can exploit these greater freedoms to produce larger perturbations, as the positions at different time steps are not linked to each other via the dynamic model. 

\subsection{Attack performance is consistent with the choice of attack objective}

Finally, we analyzed the effect of attack objective on performance measures (Table~\ref{tab:Performance_attacks_our}). Intuitively, one would expect that targeting, e.g. \acrshort{ade} would result in largest effect on the \acrshort{ade} measure compared to other attack objectives, yet understanding how other measures (including magnitude and feasibility measures) are affected can provide a better insight into implications of different attacks.

We observed that the \acrshort{fde} loss resulted in larger perturbations than the \acrshort{ade} loss, with the only exception being the mean displacement $D_{\text{mean}}$ when not constraining future perturbation and constraining observed ones in a time-specific way. This then also results in higher \acrshort{ade} and \acrshort{fde} values, as well as a higher predicted collision rate $\text{CR}_{\text{pred}}$. This is somewhat expected, as the gradient under the \acrshort{fde} loss is driven solely by the time step which generally has the largest displacement, and not averaged over previous time steps with lower displacements.

Meanwhile, the picture for the false positive collision attack is somewhat different. There, attack efficiency in terms of \acrshort{ade} and \acrshort{fde} measures is lower than under the \acrshort{fde} attack, but compared to the \acrshort{ade} attack, these measures are only lower if there is no constraint used for the perturbed future trajectory. If such a constraint is used, the false positive collision attack results in larger \acrshort{ade} and \acrshort{fde} measures compared to the \acrshort{ade} attack. 
The same relationships can be seen in the perturbation magnitude measures.

\section{Discussion}
Our results demonstrate that perturbing control actions with constraints on future positions enables the generation of realistic, feasible, yet still efficient adversarial trajectories. More generally, when testing trajectory prediction models for adversarial robustness, evaluating them just based on attack efficiency measures such as \acrshort{ade}, \acrshort{fde}, or collision rate is insufficient: attack magnitude and dynamic feasibility need to be considered.

Virtually all existing work on adversarial attacks constrains maximum displacements of the perturbed observed trajectories. However, based on our investigation of dynamic feasibility measures, the typically reported resulting drop in model performance is likely caused not only by a failure in the prediction model, but more by the fact that the given inputs lie far outside not only the training domain, but also the realm of physical feasibility, especially when position and not control actions are perturbed. Consequently, such attacks (which, in addition, could easily be detected in the real world by simple filters on control states) do not represent a viable way of evaluating the robustness of prediction models, unless some effective bounds on the control states are added.

Along this line, we have also demonstrated the importance of constraining the perturbations along the future trajectory in addition to the past observation. This has a significant influence on the generated trajectories: while resulting in perturbations of the observed trajectory more comparable to the ground truth, it ensures the consistency between the adversarial agent's past and future behavior. Without it, comparing the ground truth future $Y$ with the perturbed predictions $\widehat{\widetilde{Y}}$ would not actually give any meaningful information on the robustness of the prediction model, as the ground truth $Y$ no longer has any connection to the perturbed past $\widetilde{X}$. 

One of the key insights of our work is that keeping the adversarial trajectories realistic and dynamically feasible can dramatically reduce attack efficiency in terms of distance-based measures. While our approach is an important step towards finding efficient attack strategies for evaluating adversarial robustness of AVs' trajectory prediction models, future research needs to focus on improving efficiency of dynamically feasible, realistic, and meaningful attacks.  

Complementing our contribution of the new attack approach, our novel attack objective, the \emph{false negative attack}, demonstrated that even with a state-of-the-art prediction model, \acrshort{av}s could still be deceived into a false ``sense of safety'', potentially allowing collision to be caused by adversarial agents. Therefore, more research is required to improve robustness of prediction models towards this kind of attacks. 

Meanwhile, the standard, (i.e., \emph{false positive}) collision attack did not lead to significant increases in predicted collision rates compared to the \acrshort{ade} and \acrshort{fde} attacks, especially when we constrained the perturbations of future states. While this is somewhat explained by the specific scenario chosen, comparisons to the unperturbed predictions show that further research into this attack is likely needed to improve its potential without compromising its relation to the ground truth data.

In our experiment, the search attack~\cite{zhang2022adversarial} did not improve dynamic feasibility of the basic position-based attacks, still requiring extreme control actions inconsistent with dynamical model of an actual vehicle. We believe this was mainly caused be the fact that it only uses absolute bounds on perturbed control actions, and does not impose relative bounds. As absolute bounds are dataset-dependent, the use of the L-GAP dataset limits our conclusions here. Namely, it includes several collisions which are characterized by high accelerations and curvatures that fall outside normal driving patterns. Consequently, both for for the search attack and our approach, it would likely an improvement to use a more dynamical bound, which could be set to represent some quantile value instead, only increasing for samples where this new bound is surpassed.

Besides this limitation, there are also a number of other improvement and directions that we deem valuable as future work.
\begin{itemize}
    \item Given the observed close correlation between the perturbation magnitude and the performance measures \acrshort{ade} and \acrshort{fde}, it is worthwhile to study more closely the influence of those attack hyperparameters that could influence the perturbation magnitude -- especially the maximum allowed displacement $d_{\max}$ (equation~\eqref{eq:regularization_point}) and the bounds on control action perturbations $\delta_a$ and $\delta_{\kappa}$ (equation~\eqref{eq:control_constraints}). Additionally, the influence of the length of past observations $H$ or the number of predicted timesteps (especially on predicted collision rates) should be further studied.
    \item While our work benefited from the internal consistency of the L-GAP dataset, evaluating the generalizability of the attacks would be greatly enhanced by applying them to datasets with more diverse scenarios such as NuScenes \cite{caesar_nuscenes_2020}. 
    \item Given our refined attack, it would be beneficial to study the effectiveness of various responses to this approach, such as fine tuning models on the perturbed input data, or the use of randomized smoothing~\cite{cohen2019certified} to increase robustness. Additional, testing it against methods proposed for the detection of such attacks might be beneficial~\cite{fan2024novel}.
\end{itemize}

\section{Conclusion} 
In this paper, we introduced a novel strategy for generating dynamically feasible, realistic, and meaningful adversarial attacks on trajectory prediction models. Our approach successfully deceives the attacked state-of-the-art prediction model, resulting in decreased model accuracy while maintaining plausible behavior of the adversarial agent. Our results highlight the propensity of traditional attacks to produce trajectories which would not be executable in the real world or would not allow the adversarial agent to proceed with its initially intended maneuver. Consequently, we argue for the need of a comprehensive discussion in the field about the requirements for adversarial attack needed to make them an integral part of the evaluation of the robustness of prediction model. Our work is only the first step in this direction, and we suggest areas of future research that would be beneficial to the evaluation of robustness of autonomous vehicles.

\bibliographystyle{ACM-Reference-Format}
\bibliography{articles}

\appendix

\section{Differentiable programming} \label{app:derivatives_new}
We are given the following parts:
\begin{itemize}
    \item The original trajectories $X$ and $Y$.
    \item The control state perturbations $\delta_{U_{X_{\text{tar}}}}$ and $\delta_{U_{Y_{\text{tar}}}}$, with the resulting perturbed control states $\widetilde{U}_{X_{\text{tar}}} = U_{X_{\text{tar}}} + \delta_{U_{X_{\text{tar}}}}$ and $\widetilde{U}_{Y_{\text{tar}}} = U_{Y_{\text{tar}}} + \delta_{U_{Y_{\text{tar}}}}$
    \item The dynamical model $\Phi$, which allows us to respectively combine $X$ and $\widetilde{U}_{X_{\text{tar}}}$ as well as $Y$ and $\widetilde{U}_{Y_{\text{tar}}}$ to get the perturbed trajectories $\widetilde{X}_{\text{tar}}$ and $\widetilde{Y}_{\text{tar}}$.
    \item The prediction model $\mathds{P}_{\bm{\theta}}$, with $\widehat{\widetilde{Y}} \sim \mathds{P}_{\bm{\theta}}(\cdot|\widetilde{X}_{\text{tar}})$
    \item The adversarial loss function $L = \mathcal{L}\left(X,Y, \widetilde{X}_{\text{tar}}, \widetilde{Y}_{\text{tar}}, \widehat{\widetilde{Y}}_{\text{tar}} \right)$
\end{itemize}
We then have to find the gradients of the loss function $L$ in regard to the initial control state perturbations. 
For $\delta_{U_{X_{\text{tar}}}}$, we can determine that
\begin{equation}
\begin{aligned}
    \frac{d L}{d \delta_{U_{X_{\text{tar}}}}} & = \frac{d L}{d \widetilde{U}_{X_{\text{tar}}}} \underbrace{\frac{d \widetilde{U}_{X_{\text{tar}}}}{d \delta_{U_{X_{\text{tar}}}}}}_{=1} \\
    & = \frac{d L}{d \widetilde{X}_{\text{tar}}} \frac{d \widetilde{X}_{\text{tar}}}{d \widetilde{U}_{X_{\text{tar}}}} \\
    & = \left(\frac{\partial L}{\partial \widetilde{X}_{\text{tar}}} + \frac{\partial L}{\partial {\widetilde{Y}}_{\text{tar}}} \frac{d {\widetilde{Y}}_{\text{tar}}}{d \widetilde{X}_{\text{tar}}} + \frac{\partial L}{\partial \widehat{\widetilde{Y}}_{\text{tar}}} \frac{d \widehat{\widetilde{Y}}_{\text{tar}}}{d \widetilde{X}_{\text{tar}}}\right) \frac{d \widetilde{X}_{\text{tar}}}{d \widetilde{U}_{X_{\text{tar}}}}\, .
\end{aligned}
\end{equation}
Here, the calculation of the derivatives $\frac{\partial L}{\partial \cdot}$ depends on the definition of $\mathcal{L}$, the calculation of $\frac{d {\widetilde{Y}}_{\text{tar}}}{d \widetilde{X}_{\text{tar}}}$ and $\frac{d \widetilde{X}_{\text{tar}}}{d \widetilde{U}_{X_{\text{tar}}}}$ on the dynamical model $\Phi$, and the calculation of $\frac{d \widehat{\widetilde{Y}}_{\text{tar}}}{d \widetilde{X}_{\text{tar}}}$ on the prediction model $\mathds{P}$.
Similarly, we can do the same for $d \delta_{U_{Y_{\text{tar}}}}$:
\begin{equation}
\begin{aligned}
    \frac{d L}{d \delta_{U_{Y_{\text{tar}}}}} & = \frac{d L}{d \widetilde{U}_{Y_{\text{tar}}}}\underbrace{\frac{\widetilde{U}_{Y_{\text{tar}}}}{d \delta_{U_{Y_{\text{tar}}}}}}_{=1} \\
    & = \frac{\partial L}{\partial \widetilde{Y}_{\text{tar}}} \frac{d \widetilde{Y}_{\text{tar}}}{d \widetilde{U}_{Y_{\text{tar}}}} \, .
\end{aligned}
\end{equation}
Again, the gradients in the last line depend on the loss function $\mathcal{L}$ and dynamic model $\Phi$.

\section{Online Resources}
The Supplementary Materials (with the figures referenced in this work) can be found in our github page: \url{https://github.com/jhagenus/General-Framework-update-adversarial-Jeroen/blob/main/Supplementary_Material%20-%20Examples.pdf}.

\end{document}